\documentclass[12pt]{article}

\usepackage{amsmath,amsthm, amsfonts, amssymb, amsxtra,amsopn}
\usepackage{pgfplots}
\usepgfplotslibrary{colormaps}
\pgfplotsset{compat=1.15}
\usepackage{pgfplotstable}
\usetikzlibrary{pgfplots.statistics} 
\usepackage{filecontents}
\usepackage{graphicx}
\usepackage{multirow}
\usepackage{tabu}
\usepackage{algorithmic}
\usepackage{longtable}
\usepackage{booktabs}
\usepackage{listings}
\usepackage{cmap}
\usepackage{colortbl}
\usepackage{adjustbox}
\usepackage{epsfig}
\usepackage{xstring}

\usepackage{subfigure}

\PassOptionsToPackage{hyphens}{url}

\usepackage{hyperref}
\hypersetup{colorlinks=true,linkcolor=black,citecolor=black,urlcolor=blue,filecolor=black}
\hypersetup{pdfpagemode=UseNone,pdfstartview=}

\usepackage[tableposition=top]{caption}
\usepackage{enumitem}
\setlist[itemize]{noitemsep, topsep=0pt}

\advance\oddsidemargin by -0.45in
\advance\textwidth by 0.9in

\advance\topmargin by -0.375in
\advance\textheight by 0.75in

\long\def\symbolfootnotetext[#1]#2{\begingroup%
\def\thefootnote{\fnsymbol{footnote}}\footnotetext[#1]{#2}\endgroup}

\DeclareMathOperator{\thth}{th}

\pgfkeys{
    /pgf/number format/fixed zerofill=true }
    
\pgfplotstableset{
    color cells/.code={%
        \pgfqkeys{/color cells}{#1}%
        \pgfkeysalso{%
            postproc cell content/.code={%
                \begingroup
                \pgfkeysgetvalue{/pgfplots/table/@preprocessed cell content}\value
\ifx\value\empty
\endgroup
\else
                \pgfmathfloatparsenumber{\value}%
                \pgfmathfloattofixed{\pgfmathresult}%
                \let\value=\pgfmathresult
                \pgfplotscolormapaccess
                    [\pgfkeysvalueof{/color cells/min}:\pgfkeysvalueof{/color cells/max}]%
                    {\value}%
                    {\pgfkeysvalueof{/pgfplots/colormap name}}%
                \pgfkeysgetvalue{/pgfplots/table/@cell content}\typesetvalue
                \pgfkeysgetvalue{/color cells/textcolor}\textcolorvalue
                \toks0=\expandafter{\typesetvalue}%
                \xdef\temp{%
                    \noexpand\pgfkeysalso{%
                        @cell content={%
                            \noexpand\cellcolor[rgb]{\pgfmathresult}%
                            \noexpand\definecolor{mapped color}{rgb}{\pgfmathresult}%
                            \ifx\textcolorvalue\empty
                            \else
                                \noexpand\color{\textcolorvalue}%
                            \fi
                            \the\toks0 %
                        }%
                    }%
                }%
                \endgroup
                \temp
\fi
            }%
        }%
    }
}

\def\srowvecc#1#2{(\!\begin{array}{cc} 
      \noexpandarg\IfBeginWith{#1}{-}{\! #1}{#1}
    & #2\kern-0.5pt\end{array}\!)}
\def\rowvecc#1#2{\left(\!\begin{array}{cc} 
      \noexpandarg\IfBeginWith{#1}{-}{\! #1}{#1}
    & #2\kern-0.5pt\end{array}\!\right)}
\def\rowveccc#1#2#3{\left(\!\begin{array}{ccc} 
      \noexpandarg\IfBeginWith{#1}{-}{\! #1}{#1}
    & #2 
    & #3\kern-0.5pt\end{array}\!\right)}
\def\rowvecccc#1#2#3#4{\left(\!\begin{array}{cccc}
      \noexpandarg\IfBeginWith{#1}{-}{\! #1}{#1}
    & #2 
    & #3 
    & #4\kern-0.5pt\end{array}\!\right)}
\def\srowvecccc#1#2#3#4{\bigl(\!\begin{array}{cccc}
      \noexpandarg\IfBeginWith{#1}{-}{\! #1}{#1}
    & #2 
    & #3 
    & #4\kern-0.5pt\end{array}\!\bigr)}
\def\rowveccccc#1#2#3#4#5{\left(\!\begin{array}{ccccc} 
      \noexpandarg\IfBeginWith{#1}{-}{\! #1}{#1}
    & #2
    & #3
    & #4
    & #5\kern-0.5pt\end{array}\!\right)}
\def\srowvecccccc#1#2#3#4#5#6{(\!\begin{array}{cccccc} 
      \noexpandarg\IfBeginWith{#1}{-}{\! #1}{#1}
    & #2
    & #3
    & #4
    & #5
    & #6\kern-0.5pt\end{array}\!)}
\def\rowvecccccc#1#2#3#4#5#6{\left(\!\begin{array}{cccccc} 
      \noexpandarg\IfBeginWith{#1}{-}{\! #1}{#1}
    & #2
    & #3
    & #4
    & #5
    & #6\kern-0.5pt\end{array}\!\right)}

\pgfkeys{
    /pgf/number format/precision=4, 
    /pgf/number format/fixed zerofill=true }
    
\pgfplotstableset{
    /color cells/min/.initial=0,
    /color cells/max/.initial=1000,
    /color cells/textcolor/.initial=,
    color cells/.code={%
        \pgfqkeys{/color cells}{#1}%
        \pgfkeysalso{%
            postproc cell content/.code={%
                \begingroup
                \pgfkeysgetvalue{/pgfplots/table/@preprocessed cell content}\value
\ifx\value\empty
\endgroup
\else
                \pgfmathfloatparsenumber{\value}%
                \pgfmathfloattofixed{\pgfmathresult}%
                \let\value=\pgfmathresult
                \pgfplotscolormapaccess
                    [\pgfkeysvalueof{/color cells/min}:\pgfkeysvalueof{/color cells/max}]%
                    {\value}%
                    {\pgfkeysvalueof{/pgfplots/colormap name}}%
                \pgfkeysgetvalue{/pgfplots/table/@cell content}\typesetvalue
                \pgfkeysgetvalue{/color cells/textcolor}\textcolorvalue
                \toks0=\expandafter{\typesetvalue}%
                \xdef\temp{%
                    \noexpand\pgfkeysalso{%
                        @cell content={%
                            \noexpand\cellcolor[rgb]{\pgfmathresult}%
                            \noexpand\definecolor{mapped color}{rgb}{\pgfmathresult}%
                            \ifx\textcolorvalue\empty
                            \else
                                \noexpand\color{\textcolorvalue}%
                            \fi
                            \the\toks0 %
                        }%
                    }%
                }%
                \endgroup
                \temp
\fi
            }%
        }%
    }
}

\title{Machine Learning and Deep Learning for Fixed-Text Keystroke Dynamics}

\author{Han-Chih Chang\footnotemark[1]\ \ \ 
Jianwei Li\footnotemark[1]\ \ \ 
Ching-Seh Wu\footnotemark[1]\ \ \ 
Mark Stamp\footnotemark[1]\,\,\footnotemark[2]}

\begin{document}

\symbolfootnotetext[1]{Department of Computer Science, San Jose State University}
\symbolfootnotetext[2]{mark.stamp$@$sjsu.edu}

\maketitle

\abstract
Keystroke dynamics can be used to analyze the way that users type 
by measuring various aspects of keyboard input. Previous work has demonstrated 
the feasibility of user authentication and identification
utilizing keystroke dynamics. In this research, we consider a wide variety of 
machine learning and deep learning techniques based on fixed-text 
keystroke-derived features, 
we optimize the resulting models, and we compare our results to those 
obtained in related research. We find that models based on
extreme gradient boosting (XGBoost) and multi-layer perceptrons (MLP)
perform well in our experiments. Our best models outperform previous 
comparable research.

\section{Introduction}

Today, popular forms of biometric authentication include fingerprints 
and facial recognition. 
However, such biometric techniques do not resolve all authentication issues.
For example, studies show that the elderly are reluctant to use facial recognition 
and fingerprint recognition for authentication on mobile phones, 
while young people prefer to type instead of using other ways to 
authenticate~\cite{6845213}. Therefore, some passive biometric have recently 
emerged. In this research, we consider biometric based on keystroke dynamics.
Such techniques are applicable to the authentication problem, and can also
potentially play a role in intrusion detection.

Keystroke dynamics are derived from typing behavior. 
This approach typically relies on features 
such as the duration of keyboard events, the duration of the ``bounce,'' 
the time difference between each character, and so on. 
Such data can be collected through monitoring keyboard 
input and recording, for example, the time intervals between each keystroke. 
However, it is worth noting that a biometric based on keystroke dynamics is 
unlikely to be powerful enough to serve as a standalone authentication technique,
and hence keystroke dynamics generally must be used in 
conjunction with other types of authentication, such as passwords. 
In its related role as an IDS, keystroke dynamics may be competitive 
with other approaches.

Compared with popular biometric technologies such as fingerprints and iris scans, 
keystroke dynamics has some advantages. First, in terms of hardware, 
keystroke features can be gathered through a simple API interface, 
with the collected data then passed to a model for evaluation. 
Hence, no additional hardware deployment is involved, 
which reduces the cost. Second, as alluded to above, 
keystroke information can be obtained in a more passive and natural
manner, which eases the collection burden on users. 
Third, keystroke dynamics can be used in an ongoing, 
real-time IDS mode to judge whether current behavior is consistent 
with a specific user's previous behavior. In contrast, in a typical username 
and password authentication scenario, such passive
monitoring is not an option. Therefore, keystroke dynamics can serve to 
enhance security beyond the authentication phase. 

Of course, there are also some disadvantages to using keystroke dynamics 
for authentication. One issue is that if a user has an injured hand or is 
simply distracted or overly emotional, their typing patterns may not be 
consistent with the patterns used for training. Furthermore, 
another disadvantage is that typing patterns may vary based on 
different keyboards, or even due to new applications or software updates, 
which indicates that models must be updated regularly. Although such 
concerns are legitimate, it is clear that these issues can be mitigated, 
and hence the utilization of keystroke dynamics is likely to 
increase in the near future.

In this research, we analyze various keystroke dynamics data and train 
machine learning and deep learning models to distinguish between users. 
Features include individual key presses and flight time, among others. 
Note that for the sake of user privacy, we do not store sequences of actual 
keystrokes, and hence the text itself is not used for modeling purposes.

We consider a wide variety of learning techniques, including 
$k$-nearest neighbors ($k$-NN),
random forests,
support vector machines (SVM),
convolutional neural networks (CNN), 
recurrent neural networks (RNN), 
long short-term memory (LSTM) networks, 
extreme gradient boosting (XGBoost), 
and multilayer perceptrons (MLP). 

Much of the previous research in this field is based on multiclass models
trained on relatively small amounts of data per user. There are several 
inherent problems with such an approach. For example, if a new user is added, or the typing 
content (e.g., password) is changed, the model needs to be retrained.
Furthermore, until recently, most work in this considered only 
traditional statistical machine learning methods, 
with limited use of modern deep learning techniques.

The remainder of this paper is organized as follows. Section~\ref{Background} 
discusses relevant background topics, including a survey of previous work. 
Section~\ref{Implementation} describes the dataset used in our experiments 
and outlines the machine learning techniques that we employ. 
Our experimental results are presented in 
Section~\ref{Experiments and Results on Fixed-text Dataset}. 
Lastly, Section~\ref{Conclusion and Future Work} summarizes our 
main results and we include a discussion of possible directions 
for future work.

\section{Background}\label{Background}

In this section, we discuss keystroke dynamics in general and we
consider previous work in this field. In the next section,
we introduce the dataset and the various
machine learning models that we use in this research. 

\subsection{Keystroke Dynamics}

According to~\cite{Monrose}, ``keystroke dynamics is not what you type, 
but how you type.'' Most previous work on typing biometrics can be 
divided into either classification based on a fixed-text or 
authentication based on free-text~\cite{4}. For fixed text, the text used to 
model the typing behavior of a user and to authenticate the user is the same. 
This approach is usually applied to short text sequences, such as passwords. 
Classification can be based on various timing features 
related to the characters typed~\cite{10.1007/1-4020-8143-X_18}. 
Moreover, by combining a password along with a username, 
such a system can be further strengthened~\cite{7477228}. 
A comprehensive discussion related to the fixed-text data problem 
can be found in~\cite{4}. 

As for the free text case, the text used to model typing 
behavior of a user and to authenticate the user is not necessarily the same. 
This approach is usually applied to long text sequences, and can be viewed as 
a continuous form of authentication or as an intrusion detection system (IDS).
Again, in this paper we only consider the fixed-text problem.

Previously, many different distance-based methods have been applied
to keystroke dynamics. More recently, machine learning techniques 
have been considered, including support vector machines (SVM), 
recurrent neural networks (RNN), and so on~\cite{Zhong}. 
The learning techniques evaluated in this paper
are introduced below.

\subsection{Previous Work}

In this section, we first consider distance-based methods.
Then we discuss more recent work that relies on
various machine learning techniques.

The concept of keystroke dynamics first appeared in the 1970s 
and was focused on fixed-text data~\cite{forsen1977personal}. 
In subsequent years, Bayesian classifiers based on the mean and variance in 
time intervals between two or three consecutive key presses 
were applied to the problem~\cite{MONROSE2000351}. 
The result in~\cite{MONROSE2000351} claim a classification accuracy of~92\%\ on
a dataset with~63 users. 

Typical of early work in this field are
nearest neighbor classifiers based on various distance measures.
Initially, Euclidean distance or, equivalently, the~$L_2$ norm was used. 
In contrast to the~$L_2$ norm, the~$L_1$ norm (i.e., Manhattan distance) 
makes it easier to determine the contributions made by individual
components, and it is more robust to the effect of outliers. 
In~\cite{5}, it is shown that among all 
distance-based techniques, the best performance is obtained from 
a nearest neighbor classifier that uses a scaled Manhattan 
distance.

Neither the~$L_1$ nor the~$L_2$ norm deal effectively 
with statistical properties, and hence statistical-based distance measures
have also been considered. For example, Mahalanobis distance 
has been widely used in keystroke dynamics research~\cite{62613}.

Recently, research in keystroke dynamics has been heavily focused
on machine learning techniques. Such research 
includes $k$-nearest neighbors ($k$-NN)~\cite{5634492}, 
$K$-means clustering~\cite{10.1007/978-3-540-74549-5_125}, 
random forests~\cite{5544311}, 
fuzzy logic~\cite{886039}, 
Gaussian mixture models~\cite{4656564}, 
and many other approaches. In the remainder of this section, 
we discuss some relevant examples of machine learning based research
focused on fixed-text keystroke dynamics.

In~\cite{1223761}, support vector machines (SVM) 
are used to extract features from the data that are 
then used for classification. 
Another popular machine learning technique has been used in
keystroke dynamics is hidden Markov models (HMM).
An HMM includes a Markov process that is ``hidden'' in the sense
that it can only be indirectly observed~\cite{Stamp04arevealing}. 
In~\cite{1431505}, an HMM is used to learn the time intervals in keystroke dynamics.

A number of neural network architectures have also been applied in keystroke 
dynamics in recent years~\cite{10.1006/imms.1993.1092, 611659}. 
Deep learning techniques have also been successfully applied to 
classification and have achieved better performance, as compared to previous 
techniques, such as those considered in~\cite{MULIONO2018564}. 
Deep networks usually require a relatively long time to train, and hence 
Adam optimization and leaky rectified linear unit (leaky relu) 
activation functions are 
often used to speed up the learning process~\cite{7}. 

In~\cite{8704135}, a genetic algorithm known as
neuro evolution of augmenting topologies (NEAT) is considered.
This algorithm achieves a high accuracy on a custom dataset. 

In~\cite{1}, keystroke dynamics authentication based on
fuzzy logic is considered, and an accuracy of~98\%\ is achieved. This model evolves 
in the sense that it can update keystroke templates when a user login is successful. 
The research in~\cite{e} uses extreme gradient boosting (XGBoost), 
random forest, multilayer perceptron (MLP), and other machine learning methods to 
perform multiclass classification on 
the Carnegie Mellon University (CMU) dataset, which is the same dataset 
considered in this paper. 
In~\cite{e}, a highest accuracy of~93.79\%\ is achieved using XGBoost. 
However, these authors
do not discuss hyperparameter tuning, and thus it may
be possible to improve on their results.

As the name suggests, the equal error rate (EER) is the point where 
the false acceptance rate (FAR) and false rejection rate (FRR),
at which point the sum of the FRR and FAR is minimized. 
The value of the EER is serves as an indicator of the performance 
of a system, enabling the direct comparison of different biometrics---the 
lower the value of EER, the better the performance of 
the system. The EER is easily obtained from an ROC curve.

The authors of~\cite{8267667} propose using 
convolutional neural networks (CNN) for authentication
based on keystroke dynamics. 
Their model architecture is very similar to that in~\cite{13}, with
the main ideas deriving from a sentence classification task. They feed time-based 
feature vectors into the model directly instead of reshaping the vectors into matrices. 
They also explore the influence of different kernel sizes, different numbers of kernels, 
and different numbers of neurons in the fully connected layer. Their model is evaluated on an 
open fixed-text keystroke dataset, and their best equal error rates (EER) are 2.3\%\ and 6.5\%\ with 
and without data augmentation, respectively.

Time-based features and pressure-based features
are considered in~\cite{2}. 
By combining the information of these two kinds of features, 
the authors achieve good performance. In addition, they deal with 
typos---when a typo is recognized, the duration of keystroke time between 
the wrong key and back-space key is ignored, as is the duration between the back-space 
key and the correct key. 

Another study considers deep belief networks (DBN) to extract hidden features, 
which are then used to tune a pre-trained neural network~\cite{Belief}. 
The authors of~\cite{Belief} claim that deep learning techniques 
significantly outperform other algorithms on the CMU fixed-text dataset.  

The CMU keystroke dataset is a well known public fixed-text dataset and 
has been extensively studied. The use of a common dataset enables research
to be directly compared. In~\cite{5}, the authors introduce this dataset and achieve a 
baseline result with an EER of~9.6\%. 
There are now many studies that use this same dataset and 
outperform this baseline result. For example, in~\cite{8267667}, 
the authors obtain an EER of~2.3\%, based on a CNN with data augmentation, 
while in~\cite{7}, an EER of~3\%\ is attained 
using a multi-layers perceptron (MLP).

As an aside, we note that other keystroke features might be of interest.
For example, keystroke acoustics for user authentication are considered 
in~\cite{6966780}. In this research, a dataset containing~50 users 
results in an EER of~11\%, which shows that acoustical information 
can be informative. However, an advantage of
keystroke dynamics is that such information is easily collected 
directly from any standard keyboard.

\section{Implementation}\label{Implementation}

In this section, we first introduce the fixed-text keystroke dynamics dataset considered
in this research. Then we briefly discuss the various learning techniques 
that we have applied to this dataset.

\subsection{Dataset}\label{sect:dataset}

The Carnegie-Mellon University (CMU) 
fixed-text dataset is used
for all experiments considered in this paper.
The CMU dataset commonly serves to
benchmark techniques in keystroke dynamics
research~\cite{8883621,8267667,8628852,Gedikli,e,7,7435705}. 
This dataset includes~51 users’ keystroke dynamics information,
where each user typed the password ``\texttt{.tie5Roanl}'' a total of~400 times,
consisting of~50 repetitions over each of~8 sessions. Between sessions, 
a user had to wait at least one day, so that the day-to-day variation of 
each subject's typing was captured~\cite{5}. Furthermore this 
password was chosen to be representative of a 
strong 10-character password, as it contains a special symbol, 
a number, lowercase letters, and a capital letter.
Each time this password is typed, 31 time-based features were collected, 
as listed in Table~\ref{tab:CMUdata}. 
Note that the \texttt{Enter}
key is pressed after typing the 10-character password. 
Hence, there are~11 keystrokes, consisting of~10 consecutive pairs.

\begin{table}[!htb]
\caption{Keystroke features in CMU dataset}\label{tab:CMUdata}
\centering
\adjustbox{scale=0.85}{
\begin{tabular}{c | ccl}\midrule\midrule
Notation & Number & Summary & Description\\ \midrule
H & 11 & hold time & The length of time that a key is pressed\\[1.5ex]
DD & 10 & down-down & The length of time from one key press \\
                                 &&& to the next key press\\[1.5ex]
UD & 10 & up-down & The length of time from one key being\\
                            &&& released until the next key is pressed\\ \midrule
Total & 31 & --- & ---\\
\midrule\midrule
\end{tabular}
}
\end{table}

Individual keystrokes in a sequence can be viewed as words in a sentence. 
Based on this concept, we can tie the UD-time and DD-time from two adjacent keystrokes 
with the duration of the previous keystroke. Following this approach, 
for each keystroke, we obtain a vector 
consisting of three features, which we interpret as
an~$11 \times 3$ matrix. Thus, our feature ``vectors'' consist
of a sequence of these matrices. We refer
to this matrix as the ``fixed keystroke dynamics sequence,'' 
which we abbreviate as fixed-KDS. 

\subsection{Learning Techniques}

For our experiments, we have considered a wide variety
of learning techniques.
We introduce these learning techniques in this section.

\subsubsection{Random Forest}\label{sec:rf}

A random forest~\cite{3} is a supervised, decision tree-based machine learning method that
is often highly effective for classification and regression tasks.  This technique consists of a large 
number of individual decision trees, where each decision tree is based on a subset of 
the available features, and a subset of the training samples. The subsets 
used for each decision tree are selected with replacement. 
A majority vote or averaging of the component decision trees 
is used to determine the random forest classification.

\subsubsection{Support Vector Machine}\label{sec:svm}

Support vector machines (SVM)~\cite{4} are a powerful class 
of supervised machine learning techniques. 
The key idea of an SVM is to construct a hyperplane, so that the data can 
be divided into categories~\cite{StampML2017}. 
The so-called ``kernel trick'' enables us to 
efficiently deal with nonlinear transformations of the feature data.
As with random forests, SVMs often perform well in practice.

\subsubsection{$K$-Nearest Neighbors}\label{sec:$k$-NN}

The $k$-nearest neighbors ($k$-NN) algorithm~\cite{5} is an intuitively simple technique,
whereby we classify a sample based on the~$k$ nearest samples in the training set.
In spite of its simplicity, $k$-NN often performs well, although overfitting is a concern,
especially for small values of~$k$.
Both $k$-NN and random forest are neighborhood-based algorithms, although the neighborhood
structure determined by each is significantly different.

\subsubsection{T-SNE}\label{T-SNE}

The method of t-distributed stochastic neighbor embedding (t-SNE)
is a non-linear dimensionality reduction technique that was originally proposed in~\cite{6}. 
It is typically used for data visualization, to reduce the dimensionality of the feature space,
and for clustering. In contrast to the more well-known
principal component analysis (PCA), 
t-SNE is better able to capture non-linear relationships in the data. 

\subsubsection{XGBoost}\label{XGBoost}

XGBoost, the name of which is derived from extreme gradient boosting,
is a popular technique that has played an important role in a
large number of Kaggle competitions. In comparison to the simpler AdaBoost
technique, XGBoost has advantages in terms of dealing with outliers and
misclassifications.

Data augmentation consists of generating synthetic data based on
an existing dataset. Such ``fake'' data can be used to make up for a lack of data
for a given problem. Data augmentation has often proved valuable in practice.
We consider data augmentation in our XGBoost experiments.

\subsubsection{LSTM and Bi-LSTM}\label{LSTM}

Long short-term memory (LSTM) is a highly specialized
recurrent neural network (RNN) architecture  that is able to better 
deal with the vanishing and exploding gradient issues that plague 
plain ``vanilla'' RNNs~\cite{StampML2017}. 
Consequently, LSTMs generally perform much better over longer sequences
as compared to vanilla RNNs.

A bi-directional LSTM (bi-LSTM) combines two LSTMs, one computed in the
forward direction and another computed in the backward direction. 
Bi-LSTMs are well-suited to sequence labeling tasks and have
proven to be strong at modeling contextual information in 
natural language processing (NLP) tasks.

In our LSTM and bi-LSTM experiments, we consider two different encoding
methods. In addition to the standard raw feature encoding, we also 
experiment with one-hot encoding. Assuming that a feature can take on~$m$ 
possible values, a feature value of~$k$ has a one-hot representation
consisting of a binary vector of 
length~$m$ with a~1 in the~$k^{\thth}$ position and~0 elsewhere.
When training, one-hot encoding has a natural interpretation as 
a vector of probabilities, and hence it is well suited to training involving 
a softmax output layer, for example.

In out LSTM and bi-LSTM experiments, we also consider attention mechanisms.
The idea of an attention mechanism is intuitively simple---we want to force the 
model to focus on some specific aspect of the training data.
Attention is somewhat related to regularization, in the sense that we
reduce the potential for over-reliance on some parts of the training
data, which can lead to various pathologies, including overfitting.

\subsubsection{Convolutional Neural Network}\label{CNN}

Convolutional neural networks (CNN) are designed to deal effectively 
and efficiently with local structure. CNNs have proven their 
worth in the realm of image analysis. Most CNN architectures include
convolutional layers, pooling layers, and a fully-connected output layer. 

\subsubsection{Multi-Layer Perceptron}\label{MLP}

The structure of a generic multi-layer perceptron (MLP) 
includes an input layer, one or more hidden layers, 
and an output layer. Each node, or neuron, in a hidden layer
includes a nonlinear activation function, which is the key to the ability of
an MLP to deal with challenging data. To mitigate overfitting, 
we employ dropouts for regularization in our MLP experiments~\cite{7}.

\section{Experiments and Results}
\label{Experiments and Results on Fixed-text Dataset}

This section contains the results of our fixed-text experiments on the CMU dataset.
We provide some analysis and discussion of our results.

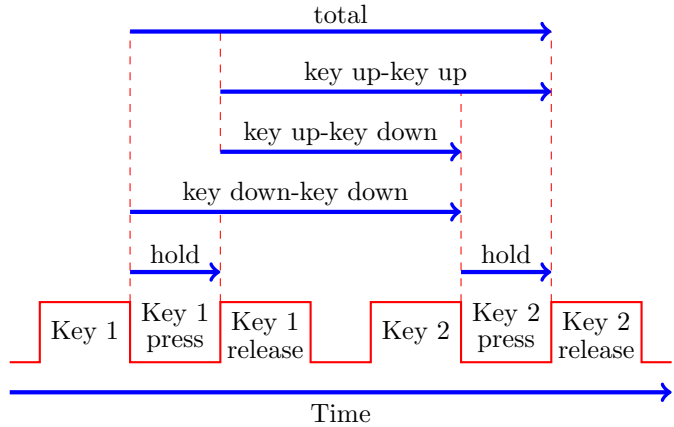
\begin{figure}[!htb]
\centering
    \begin{tikzpicture}[scale=0.8]
    
    \draw[red,thick] plot[] coordinates {
    (-0.5,0.5) (0.0,0.5) (0.0,1.5) (1.5,1.5) (1.5,0.5) (3.0,0.5) (3.0,1.5) (4.5,1.5) (4.5,0.5)
    (5.5,0.5) (5.5,1.5) (7.0,1.5) (7.0,0.5) (8.5,0.5) (8.5,1.5) (10.0,1.5) (10.0,0.5) (10.5,0.5)};

    \draw[red,dashed] (1.5,1.5) -- (1.5,6.0);
    \draw[red,dashed] (3.0,1.5) -- (3.0,3.0);
    \draw[red,dashed] (3.0,4.0) -- (3.0,6.0);
    
    \draw[red,dashed] (7.0,1.5) -- (7.0,5.0);
    \draw[red,dashed] (8.5,1.5) -- (8.5,6.0);
    
    \node at (0.75,1.0) {\footnotesize Key~1};
    \node at (2.25,1.3) {\footnotesize Key~1};
    \node at (2.25,0.8) {\footnotesize press};
    \node at (3.75,1.15) {\footnotesize Key~1};
    \node at (3.75,0.725) {\footnotesize release};

    \node at (6.25,1.0) {\footnotesize Key~2};
    \node at (7.75,1.3) {\footnotesize Key~2};
    \node at (7.75,0.8) {\footnotesize press};
    \node at (9.25,1.15) {\footnotesize Key~2};
    \node at (9.25,0.725) {\footnotesize release};

    \draw[ultra thick,color=blue,->] (3.0,5.0) -- (8.5,5.0); 
    \node at (5.75,5.30) {\footnotesize key up-key up};

    \draw[ultra thick,color=blue,->] (3.0,4.0) -- (7.0,4.0); 
    \node at (5.0,4.30) {\footnotesize key up-key down};

    \draw[ultra thick,color=blue,->] (1.5,3.0) -- (7.0,3.0); 
    \node at (4.25,3.30) {\footnotesize key down-key down};

    \draw[ultra thick,color=blue,->] (1.5,2.0) -- (3.0,2.0); 
    \node at (2.25,2.30) {\footnotesize hold};

    \draw[ultra thick,color=blue,->] (7.0,2.0) -- (8.5,2.0); 
    \node at (7.75,2.30) {\footnotesize hold};

    \draw[ultra thick,color=blue,->] (1.5,6.0) -- (8.5,6.0); 
    \node at (5.0,6.3) {\footnotesize total};

    \draw[ultra thick,color=blue,->] (-0.5,0.0) -- (10.5,0.0); 
    \node at (5.0,-0.35) {\footnotesize Time};

     
    \end{tikzpicture}
\caption{Keystroke dynamics features}\label{fig:Keystroke-Dynamics-Features}
\end{figure}

As mentioned above, in the CMU dataset, the data is arranged as a table with~31 columns,
representing the collected information for one timing of the password.
For example, one column is \texttt{H.period} which is
the hold time for the ``\texttt{.}'' key. The hold time is the length of time when the key 
was depressed. Another example is the column \texttt{DD.period.t},
is the time interval between when the ``\texttt{.}'' key was pressed until the
``\texttt{t}'' key was pressed.
The overall table is~$20400 \times 31$, where each row corresponds to 
the timing information for a single repetition of the password by a single subject. 
Figure~\ref{fig:Keystroke-Dynamics-Features} illustrates the timing relationship between
consecutive keystrokes.

\subsection{Data Exploration}

There are~31 timing features in the CMU dataset, which can be divided
into three groups which we denote as~DD, UD and~H. Here,
we analyze the data to determine whether there is any significant difference among these 
three groups. For this data exploration, we have randomly selected six 
of the~51 subjects for analysis.

\begin{figure}[!htb]
\centering
\begin{tabular}{cc}
\includegraphics[width=0.4\textwidth]{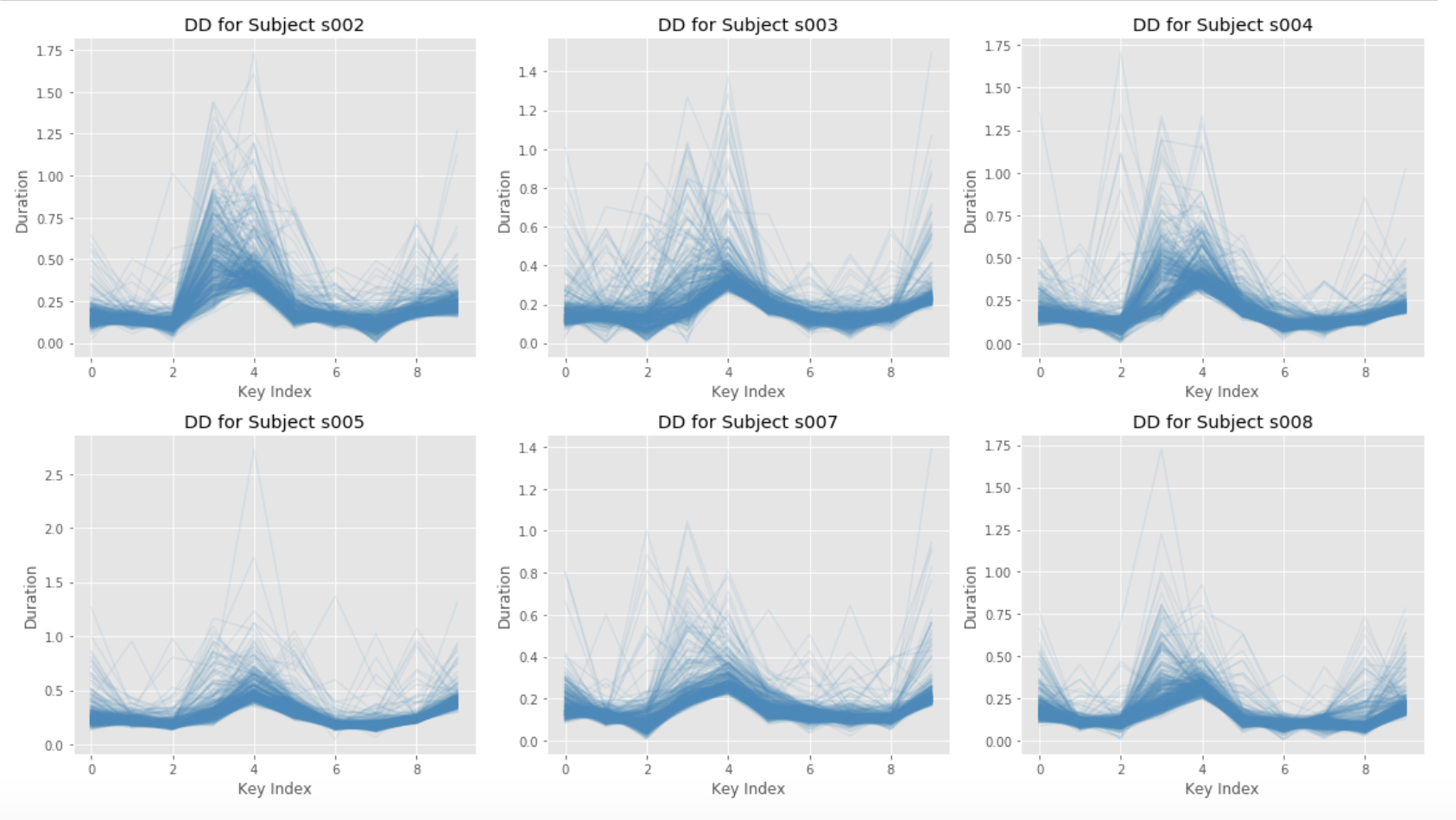}
&
\includegraphics[width=0.4\textwidth]{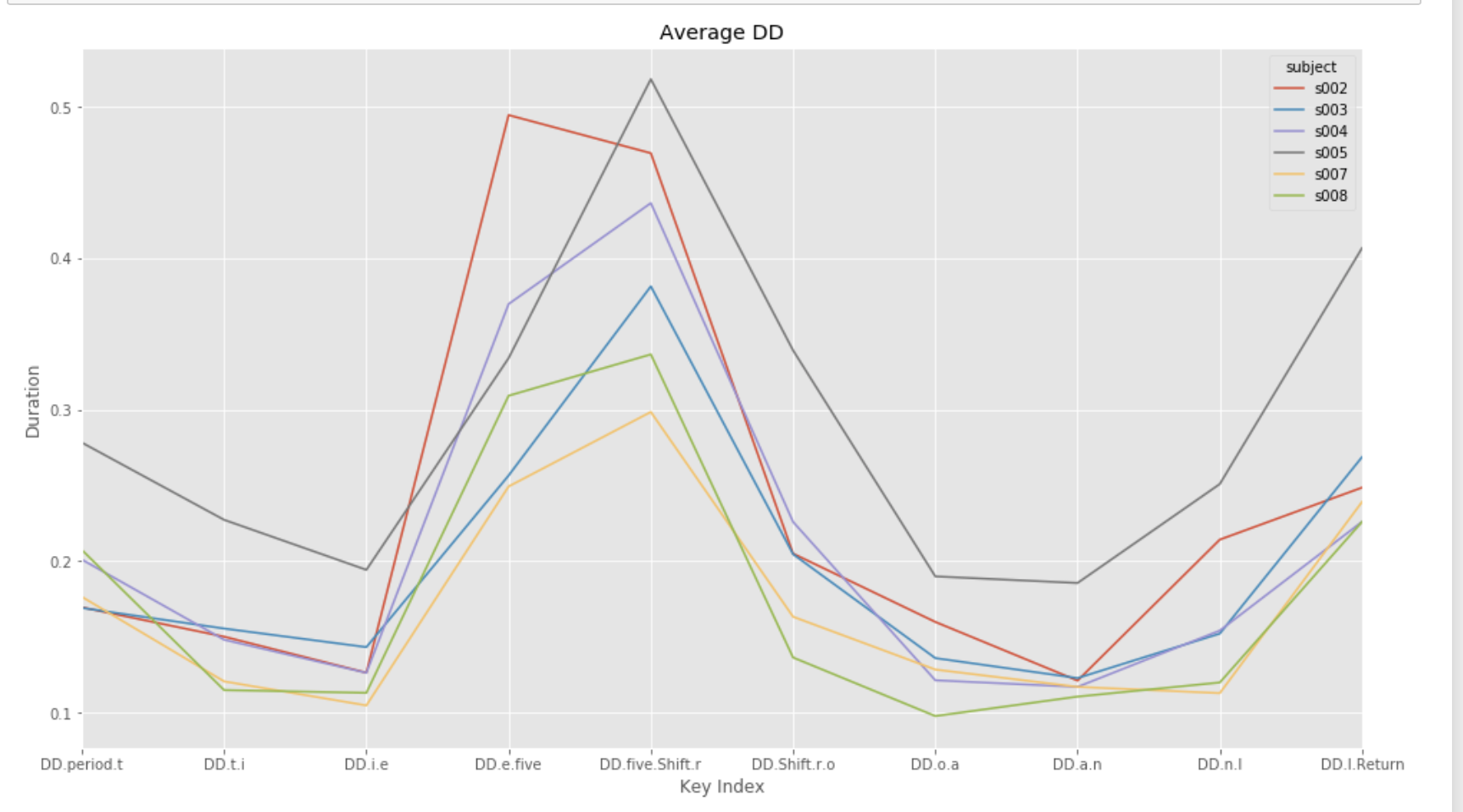}
\\
(a) Individual
&
(b) Average
\end{tabular}
\caption{Key-down key-down for six subjects (400 keystrokes)} 
\label{fig:DD-users-average}
\end{figure}

In Figure~\ref{fig:DD-users-average}~(a), each line graph represents the~400 
input feature vectors corresponding to a given subject. From this figure, we 
observe that most of the feature vectors are fairly consistent in that they
follow a similar pattern for a given subject. 
This indicates that subjects tend to be relatively consistent with respect to this
particular feature group. This observation can be seen as a positive indicator of the potential
to successfully classify the subjects. However, when the six subjects' average
cases are compared in Figure~\ref{fig:DD-users-average}~(b), 
the results show that the subjects have somewhat similar typing patterns. 

\begin{figure}[!htb]
\centering
\begin{tabular}{cc}
\includegraphics[width=0.4\textwidth]{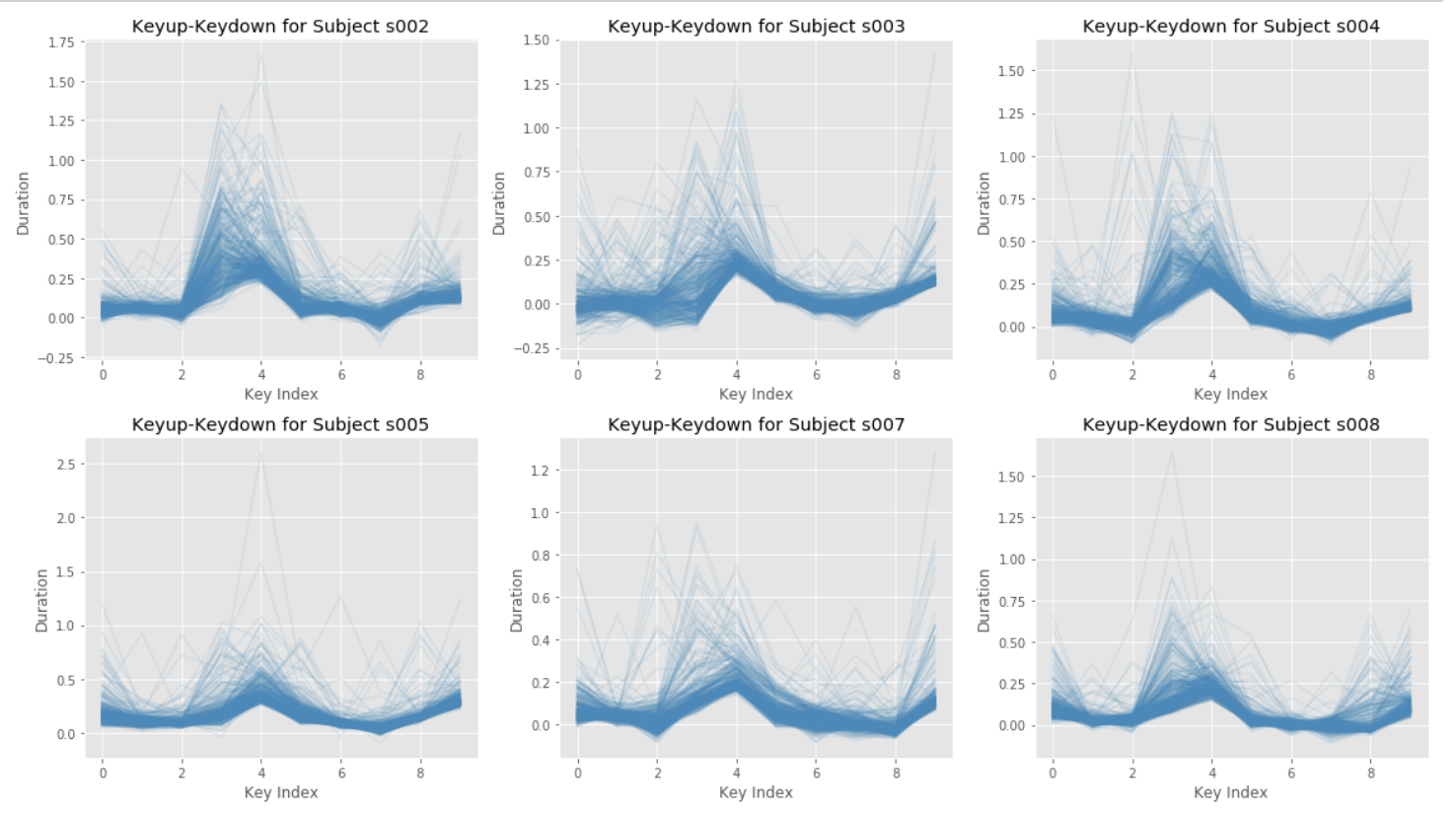}
&
\includegraphics[width=0.4\textwidth]{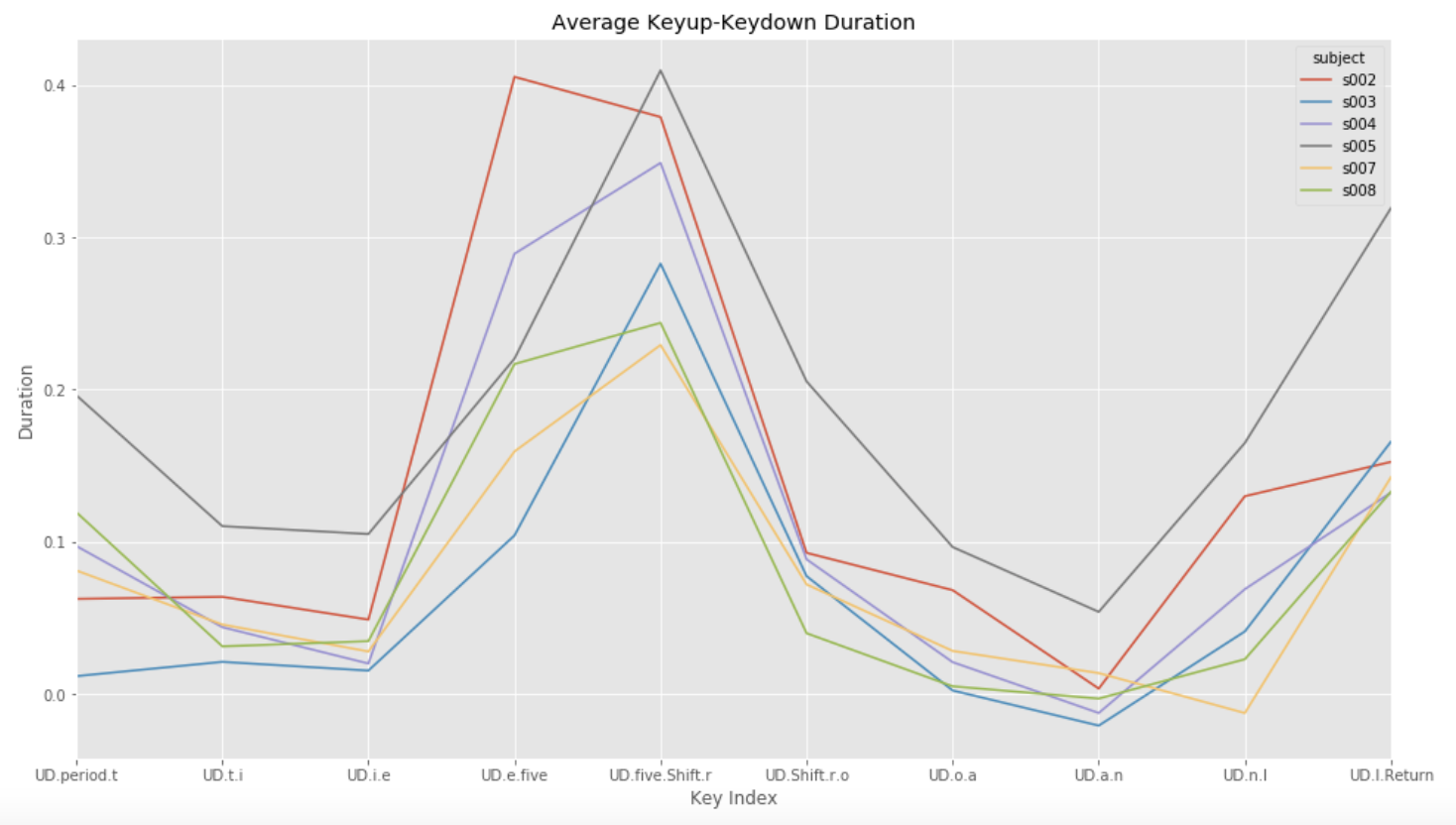}
\\
(a) Individual
&
(b) Average
\end{tabular}
\caption{Key-up key-down for six subjects for (400 keystrokes)} 
\label{fig:UD-users-average}
\end{figure}

The analogous results for the key-up key-down features
are shown in Figure~\ref{fig:UD-users-average}. 
We observe that this data is
similar to key-down key-down data
in Figure~\ref{fig:DD-users-average}.

\begin{figure}[!htb]
\centering
\begin{tabular}{cc}
\includegraphics[width=0.4\textwidth]{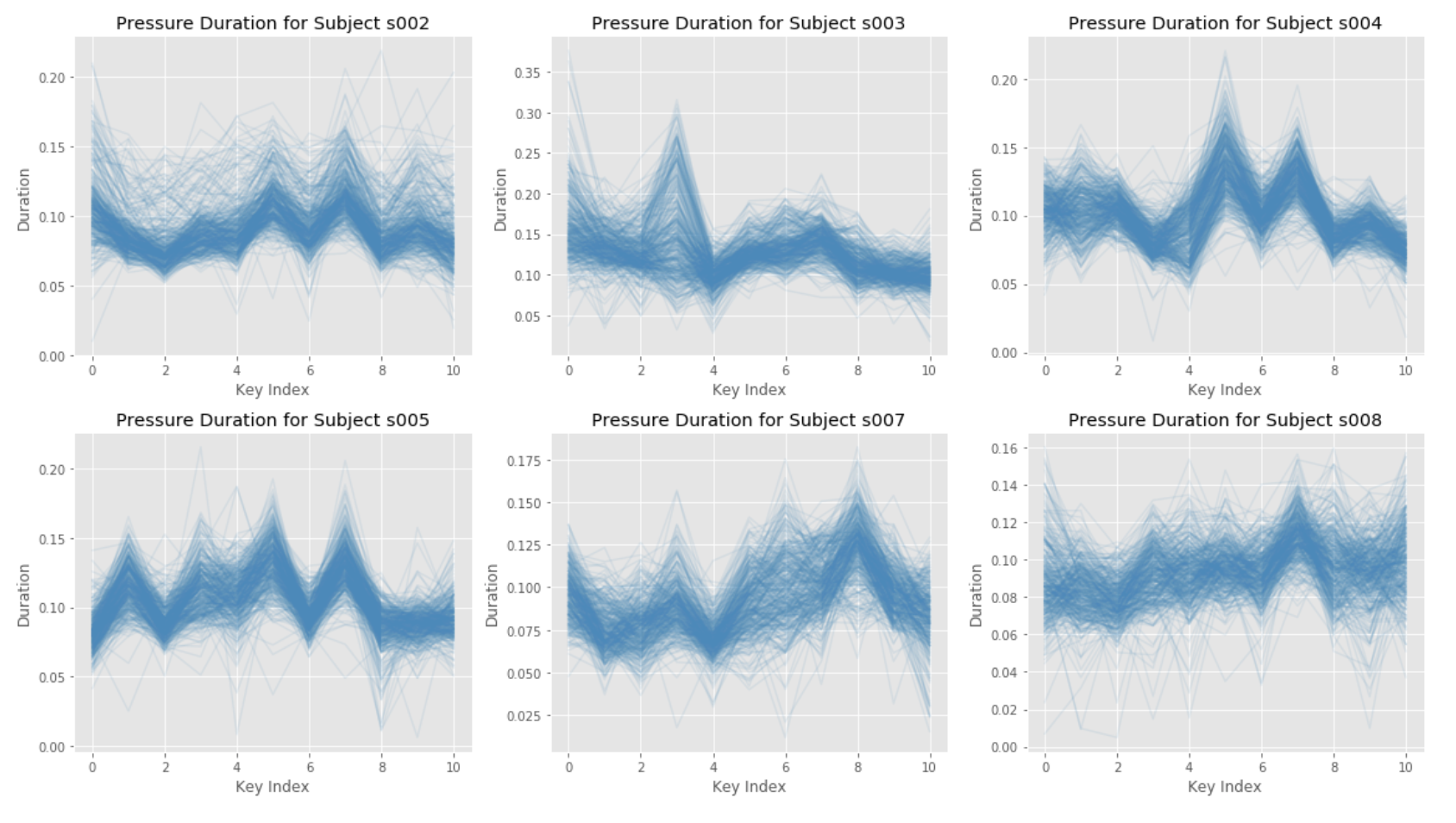}
&
\includegraphics[width=0.4\textwidth]{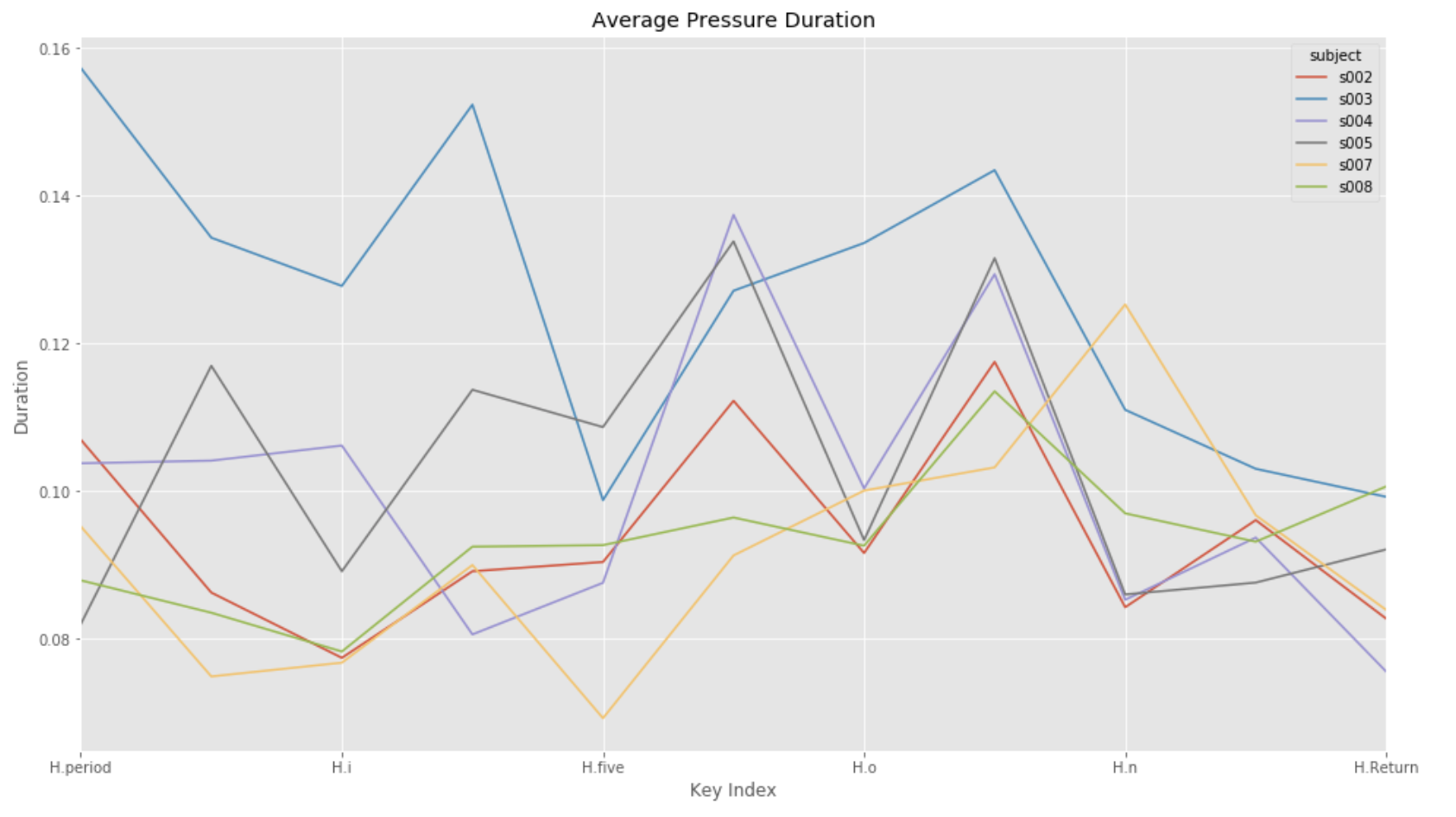}
\\
(a) Individual
&
(b) Average
\end{tabular}
\caption{Hold time for six subjects (400 iterations)} 
\label{fig:H-users-average}
\end{figure}

In Figure~\ref{fig:H-users-average}~(a), we compare the six subjects 
based on the hold-time feature, and here the differences are more pronounced.
In particular, the average cases in Figure~\ref{fig:H-users-average}~(b)
reveal more substantial differences. These results indicate that the hold duration 
should be a strong feature for distinguishing users.

\begin{figure}[!htb]
\centering
\includegraphics[width=0.5\textwidth]{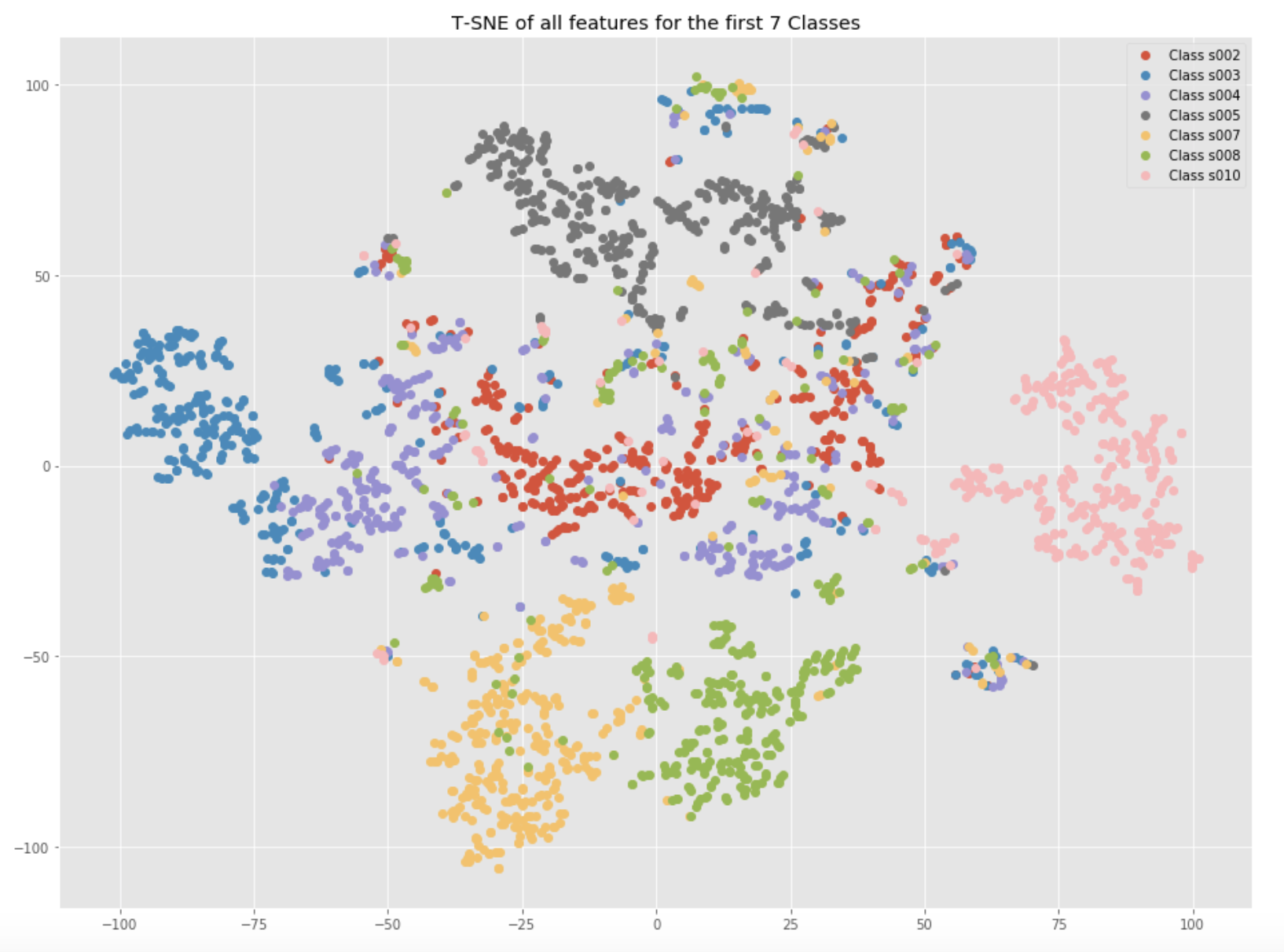}
\caption{T-SNE of features of seven subjects} 
\label{fig:tsne}
\end{figure}

To further explore the data, 
we apply t-SNE as a clustering technique to
gain insight into how the data is distributed. 
In this case, we consider a subset consisting of the first seven subjects, 
using all~400 records for each of these subjects. The result in Figure~\ref{fig:tsne}
show that the subjects can be clustered into different groups.
This is again promising, as it indicates that we should
have success in distinguishing users.

\subsection{Classification Results}

In this section, we give our classification results. Here, we 
experiment with $k$-NN, random forest, SVM, XGBoost, MLP, CNN,
RNN, and LSTM.

\subsubsection{$K$-Nearest Neighbor Experiments}

We optimize with respect to three parameters of the $k$-NN algorithm,
namely, the number of neighbors, 
the weight function used for prediction, and the distance category. 
As in all of our parameter tuning experiments,
we employ a Bayes model to generate a suit of parameters with 
the highest probability being the best result. Table~\ref{tab:3.3} 
shows the search space for each parameter and the best accuracy achieved.
Boldface entries in Table~\ref{tab:3.3} are used to indicate the optimal parameter
values.

\begin{table}[!htb]
\caption{Results for $k$-NN}\label{tab:3.3}
\centering
\adjustbox{scale=0.85}{
\begin{tabular}{c|c|c}\midrule\midrule
Parameter & Search space & Accuracy\\ \midrule
\texttt{n\_neighbors} & [\textbf{5},50] & \multirow{4}{*}{82.27\%}\\
\texttt{weight} & uniform, \textbf{distance}\\
$p$ & [\textbf{1}, 2, 3] \\
Experiments & 50 \\ \midrule\midrule
\end{tabular}
}
\end{table}

\subsubsection{Random Forest Experiments}

We optimize four parameters of the random forest algorithm, namely,  
the number of decision trees, the maximum depth of each decision tree, 
the minimal number of samples in a leaf node, 
and the minimum number of samples required to split. 
Again, we make use of different combinations of values of these parameters
to build a Bayes model, which generates
a set of parameters that will, with high probability, yield the best result.
Table~\ref{tab:3.1} shows the range considered for each of these parameters,
the optimal value that we use, and the best result obtained.

\begin{table}[!htb]
\caption{Results for random forest}\label{tab:3.1}
\centering
\adjustbox{scale=0.85}{
\begin{tabular}{c|c|c}\midrule\midrule
Parameter & Search space & Accuracy\\ \midrule
\texttt{n\_estimaters} & [100, \textbf{1000}] & \multirow{5}{*}{93.55\%} \\
\texttt{max\_depth} & None, 5, 10, 15, 20, 30, \textbf{35}, 40 \\
\texttt{min\_samples\_leaf} & [\textbf{1}, 10] \\
\texttt{min\_samples\_split} & [\textbf{2}, 5] \\ 
Experiments & 50 \\   \midrule\midrule
\end{tabular}
}
\end{table}

\subsubsection{Support Vector Machine Experiments}

Here, we consider four parameters of an SVM, namely,
the value of the regularization parameter, the kernel function, 
and the two coefficients of the kernel function. Again, a Bayes model is 
built to search the optimal values of these parameters. 
The search space for each parameter, the optimal 
values, and the best accuracy are given in Table~\ref{tab:3.2}.

\begin{table}[!htb]
\caption{Results for SVM}\label{tab:3.2}
\centering
\adjustbox{scale=0.85}{
\begin{tabular}{c|c|c|c}\midrule\midrule
Parameter & Search space & Best value & Accuracy \\ \midrule
$C$ & Real(1e-6, 1e+6, log-uniform) & 920319 & \multirow{5}{*}{88.02\%} \\
gamma & Real(1e-6, 1e+1, log-uniform) & 0.61620 \\
degree & [1, 8] & 8 \\
kernel &  linear, poly, rbf & rbf \\ 
Experiments & --- & 50 \\ \midrule\midrule
\end{tabular}
}
\end{table}

\subsubsection{XBGoost Experiments}

Next, we classify the samples using XGBoost. Here, we consider 
each of the three feature groups (DD, UD, and H) 
individually, as well as the combination of all three. 
The multi-classification results for the~51 subjects are shown 
in Table~\ref{tab:2} and the model parameters used to
achieve these results are given in Table~\ref{tab:2p}. 

\begin{table}[!htb]
\caption{Accuracy of four features for XGBoost}\label{tab:2}
\centering
\adjustbox{scale=0.85}{
\begin{tabular}{c|lc}\midrule\midrule
Feature & \multicolumn{1}{c}{Description} & Accuracy\\ \midrule
H & Hold time & 76.91\% \\
DD & Key-down Key-down & 76.39\% \\
UD & Key-up Key-down & 81.10\% \\ \midrule
All & H, DD and UD & 95.15\% \\ \midrule\midrule
\end{tabular}
}
\end{table}

\begin{table}[!htb]
\caption{Selected parameters for XGBoost}\label{tab:2p}
\centering
\adjustbox{scale=0.85}{
\begin{tabular}{c|c}\midrule\midrule
Parameters & \multicolumn{1}{c}{Value} \\ \midrule
\texttt{learning-rate} & 0.21  \\
\texttt{n-estimators} & 1000  \\
\texttt{max-depth} & 2 \\
\texttt{min-child-weight} & 1.4 \\ \midrule\midrule
\end{tabular}
}
\end{table}

Based on these results, we conduct further experiments with XGBoost. 
Given the fairly limited size of the training data, we apply a simple
data augmentation strategy---we randomly perturb each
timing feature based on a range of~$(-0.02,0.02)$.
In this experiment, we set the augmentation ratio to two, 
meaning that the amount of augmented date is two times the 
amount of original data.
We find that this data augmentation provides a slightly improvement
in the accuracy, as shown in Table~\ref{tab:3}.

\begin{table}[!htb]
\caption{Results for XGBoost}\label{tab:3}
\centering
\adjustbox{scale=0.85}{
\begin{tabular}{c|cc}\midrule\midrule
Description & \multicolumn{1}{c}{Data size} & Accuracy\\ \midrule
No augmentation & 16,320 & 95.42\% \\
Augmentation & 48,960 & 96.39\% \\ \midrule\midrule
\end{tabular}
}
\end{table}

\subsubsection{Multilayer Perceptron Experiments}

Our generic MLP consists of four fully connected layers, 
in which the number of neurons are~512, 256, 144, and~51, 
respectively. The output of the last layer is fed into a softmax function to 
calculate the corresponding probability for each class. A 
rectified linear unit (relu) 
activation function and a batch normalization layer are used in the first and second 
dense layers. We use the cross entropy loss function for this model---additional
parameters are listed in Table~\ref{tab:mul-mlp}.
This MLP model yields an impressive accuracy ot~95.96\%.

\begin{table}[!htb]
\caption{Results for MLP}\label{tab:mul-mlp}
\centering
\adjustbox{scale=0.85}{
\begin{tabular}{c|cccc|c}\midrule\midrule
\multirow{2}{*}{Model} & \multicolumn{4}{c|}{Parameters} & \multirow{2}{*}{Accuracy} \\
 & input-channel & output-channel & num-layers & learning-rate \\ \midrule
MLP         & 31 & 100 & 3 & 0.001 & 95.96\% \\ \midrule\midrule
\end{tabular}
}
\end{table}

\subsubsection{Convolutional Neural Network Experiments}

The input for our CNN model is the fixed-KDS data structure, which we
discussed in Section~\ref{Implementation}. The architecture of our
CNN is based on that of the so-called textCNN in~\cite{13}, 
which is used to process sequential text data. 
The key idea is to apply multiple rectangular kernels, instead of 
more typical square kernels. Specifically, the width of all kernels is 
the same as the embedding size for each word, so the output for each 
convolution is a one-dimension vector. Then multiple max-pooling layers 
are used to process these vectors to yield one feature for each kernel. 
Finally, these generated features are concatenated into a one-dimension vector, 
and multiple fully-connected layers are used to produce the class prediction.

\begin{figure}[!htb]
\centering
\includegraphics[width=0.75\textwidth]{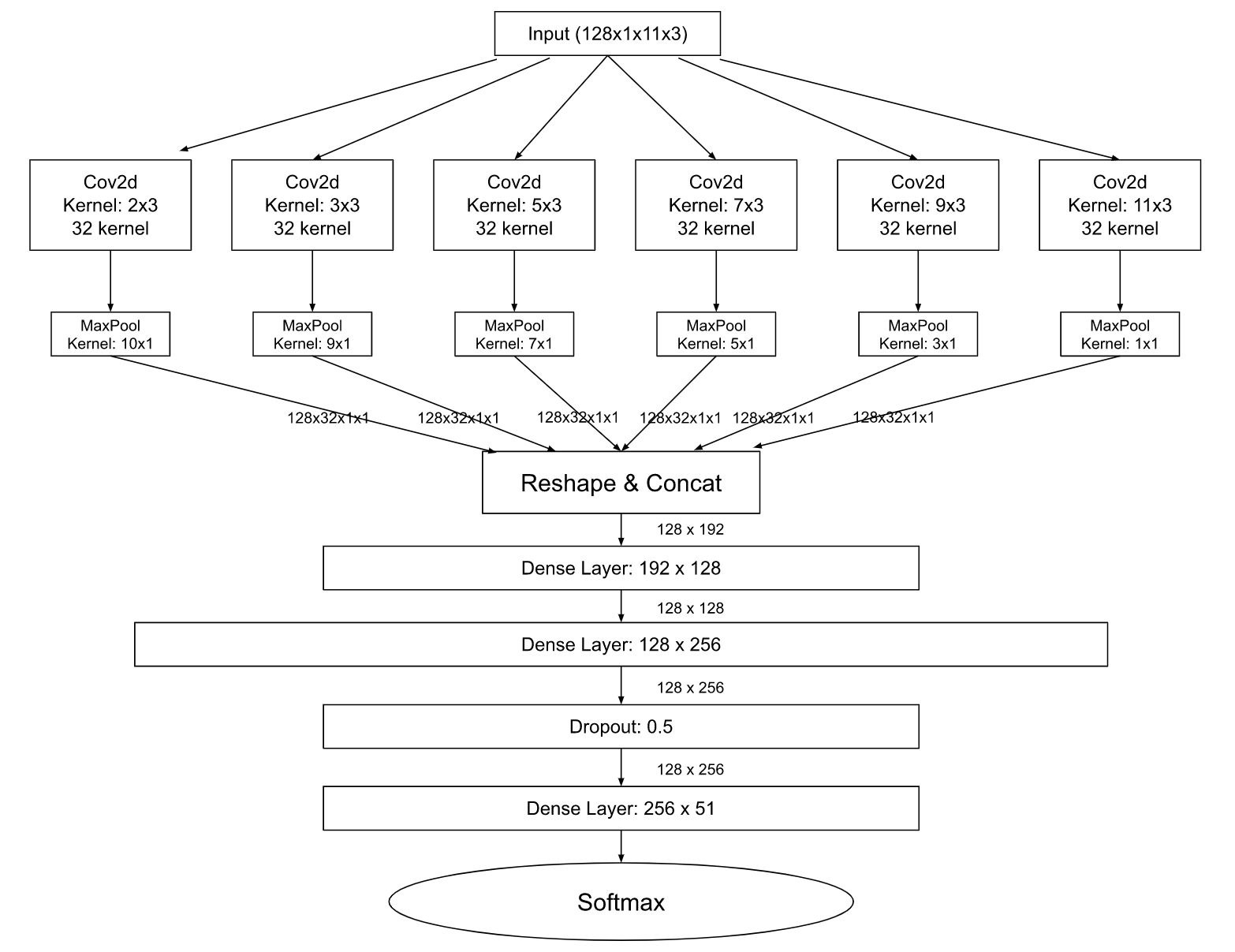}
\caption{Architecture of CNN for free-text datasets}\label{fig:cnn_architecture}
\end{figure}

In our keystroke dynamics model, we view each keystroke event as a ``word'' 
and each keystroke sequence as a ``sentence.'' In this way, six different 
convolution kernels are applied to this sequential data, and continuous max-pooling 
layers extract the most important feature from each kernel. Then the concatenated 
vector is fed into three dense layers, with a softmax function is used to generate the 
probability for each class. In addition, a dropout layer is added after the 
penultimate layer. The cross entropy loss function is used. For these 
CNN experiments, the best result we obtain is an accuracy of~92.57\%.

\subsubsection{Recurrent Neural Network Experiments}

\begin{figure}[!htb]
\centering
\includegraphics[width=0.65\textwidth]{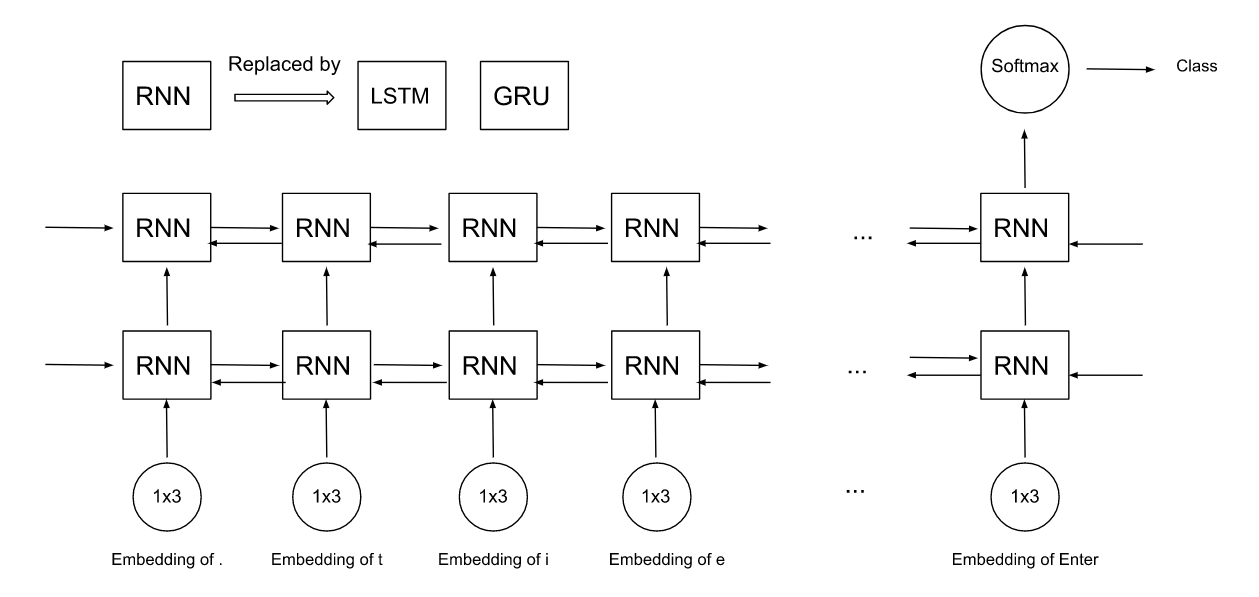}
\caption{Architecture of bi-RNN}\label{fig:rnn_architecture}
\end{figure}

The architecture of our RNN-based neural network is shown in 
Figure~\ref{fig:rnn_architecture}. The input data for this model is the fixed-KDS,
as discussed in Section~\ref{sect:dataset}. The idea behind this model comes from the 
field of sentiment analysis. Since keystroke data is inherently sequential, 
we applying a two-layers bi-directional RNN. 
In this experiment, the cross entropy loss function is used,
and the best result we obtain is an accuracy of~93.45\%.

\subsubsection{LSTM Experiments}

Next, we apply both LSTM and bi-LSTM with one-hot encoding. 
In these experiments, one-hot encoding is applied on both the subject
and the timing features, which then serve as the feature vectors for the LSTM and bi-LSTM. 
The accuracies we obtain for LSTM and bi-LSTM are shown in Table~\ref{tab:lstm}.
Although these results are reasonably strong, 
they are not competitive with our XGBoost experiments.

\begin{table}[!htb]
\caption{Results for LSTM and bi-LSTM with one-hot encoding}\label{tab:lstm}
\centering
\adjustbox{scale=0.85}{
\begin{tabular}{c|cccc|c}\midrule\midrule
\multirow{2}{*}{Model} & \multicolumn{4}{c|}{Parameters} & \multirow{2}{*}{Accuracy} \\
 & input-size & hidden-size & num-layers & learning-rate \\ \midrule
LSTM    & 31 & 5 & 1 & 0.3 & 91.28\% \\ 
Bi-LSTM & 31 & 5 & 1 & 0.3 & 90.02\% \\ \midrule\midrule
\end{tabular}
}
\end{table}

\begin{figure}[!htb]
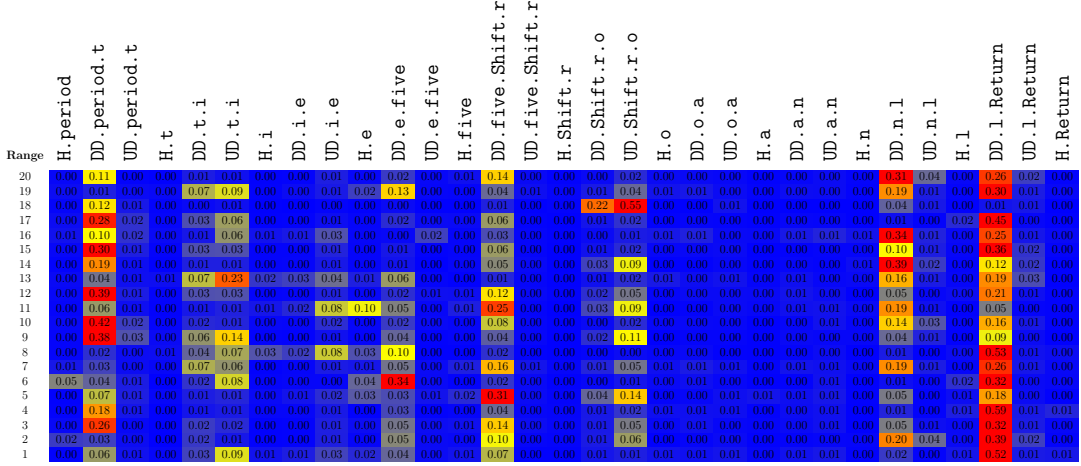

\centering
\input figures/heatmap2.tex
\caption{Attention matrix heatmap}\label{fig:attention}
\end{figure}

We further consider a bi-LSTM with attention,
primarily as a way of analyzing feature importance.
The attention matrix in the form of a heatmap, appears in Figure~\ref{fig:attention}.
This matrix
consists of the weights determined by the attention layer. 
In this matrix, the $x$-axis represents the~31 features, 
while the $y$-axis is based on~20 consecutive training samples. We observe that after several 
epochs, the attention seems to have a tendency to converge to specific 
features---at the end of the training, we find the most significant features are
\texttt{DD.period.t}, \texttt{DD.e.five}, \texttt{UD.Shift.r.o}, \texttt{DD.n.l}, and \texttt{DD.l.Return}.

\subsection{Summary and Discussion}

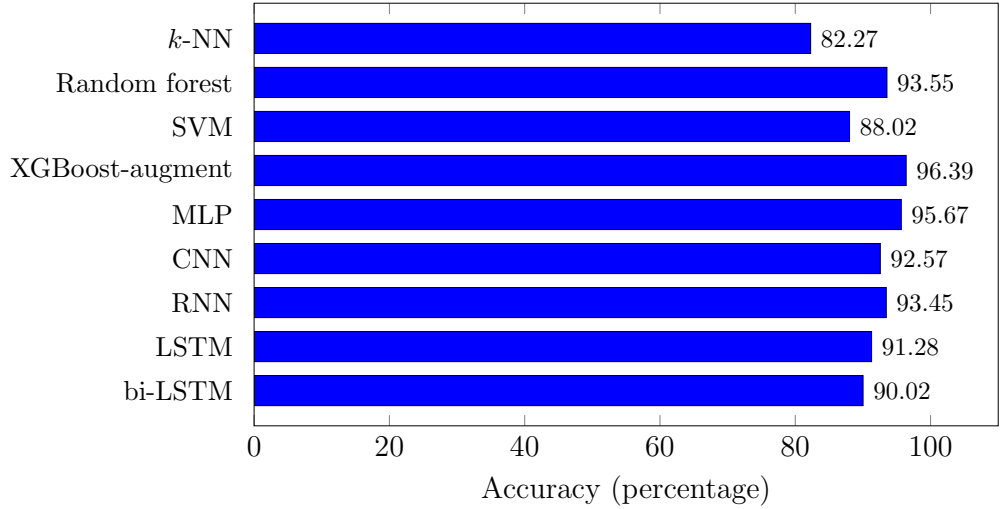
\begin{figure}[!htb]
\centering
\begin{tikzpicture}[scale=0.95, every node/.style={scale=1.0}]
\begin{axis}[ 
xbar,
width  = 0.75*\textwidth,
height = 7.5cm,
xmin=0,xmax=110.0,
xtick={0,20,40,60,80,100},
xlabel = {Accuracy (percentage)},
major y tick style = transparent,
xbar=5*\pgflinewidth,
bar width=12.0pt,
x tick label style={
    	/pgf/number format/.cd,
   	fixed,
   	fixed zerofill,
    	precision=0},
y tick label style={
		font=\small,
		},
symbolic y coords={%
	bi-LSTM,
	LSTM,
	RNN,
	CNN,
	MLP,
	XGBoost-augment,
	SVM,
	Random forest,
	$k$-NN},
ytick=data,
enlarge y limits=0.1,
nodes near coords, 
every node near coord/.append style={
	font=\footnotesize,
	/pgf/number format/.cd,
	fixed,
	fixed zerofill,
	precision=2},
ytick=data,
]
\addplot [fill=blue,opacity=1.00]
coordinates {
(90.02,bi-LSTM)
(91.28,LSTM)
(93.45,RNN)
(92.57,CNN)
(95.67,MLP)
(96.39,XGBoost-augment)
(88.02,SVM)
(93.55,Random forest)
(82.27,$k$-NN)
};
\end{axis}
\end{tikzpicture}
\caption{Summary of our results}\label{fig:overallfix}
\end{figure}

We summarize our experimental results for the CMU fixed-text dataset 
in Figure~\ref{fig:overallfix}. The result shows that among all models we have
considered, XGBoost with data augmentation, denoted XGBoost-augment,
achieves the highest accuracy at~96.39\%.

While XGBoost with data augmentation achieves the best results,
MLP does nearly as well. 
When comparing the training times for these two models, 
we find that XGBoost with data augmentation take~18 minutes to train, 
while MLP requires about half an hour. Both of these are very reasonable
training times, but if great efficiency during training is required, 
then XGBoost with data augmentation may be the better choice.

\begin{figure}[!htb]
\centering
\begin{tikzpicture}[scale=0.9, every node/.style={scale=1.0}]
\begin{axis}[ 
xbar,
width  = 0.75*\textwidth,
height = 7.5cm,
xmin=0,xmax=107.5,
xtick={0,20,40,60,80,100},
xlabel = {Accuracy (percentage)},
major y tick style = transparent,
bar shift=0pt,
bar width=12.5pt,
x tick label style={
    	/pgf/number format/.cd,
   	fixed,
   	fixed zerofill,
    	precision=0},
y tick label style={
		font=\small,
		},
symbolic y coords={%
	Ours,
	ELM Ravindran,
	XGBoost Dwivedi,
	CNN Dwivedi,
	SVM/NEAT Baynath,
	MLP Gedikli,
	CNN Ceker,
	MLP Maheshwary,
	XGBoost Krishna
	},
yticklabels={%
	XGBoost-augment (our research),
	ELM {(Ravindran, et al.~\cite{7435705})},
	XGBoost {(Dwivedi, et al.~\cite{8628852})},
	CNN {(Dwivedi, et al.~\cite{8628852})},
	SVM/NEAT {(Baynath, et al.~\cite{8883621})},
	MLP {(Gedikli, et al.~\cite{Gedikli})},
	CNN {(\c{C}eker, et al.~\cite{8267667})},
	MLP {(Maheshwary, et al.~\cite{7})},
	XGBoost {(Krishna, et al.~\cite{e})}
	},
ytick=data,
enlarge y limits=0.1,
nodes near coords={\pgfmathfloatifflags{\pgfplotspointmeta}{0}{}{\pgfmathprintnumber{\pgfplotspointmeta}}},
every node near coord/.append style={
	font=\footnotesize,
	/pgf/number format/.cd,
	fixed,
	fixed zerofill,
	precision=2},
ytick=data,
]
\addplot [fill=red,opacity=1.00]
coordinates {
(96.39,Ours)
(0,ELM Ravindran)
(0,XGBoost Dwivedi)
(0,CNN Dwivedi)
(0,SVM/NEAT Baynath)
(0,MLP Gedikli)
(0,CNN Ceker)
(0,MLP Maheshwary)
(0,XGBoost Krishna)
};
\addplot [fill=blue,opacity=1.00]
coordinates {
(0,Ours)
(89.27,ELM Ravindran)
(0,XGBoost Dwivedi)
(0,CNN Dwivedi)
(0,SVM/NEAT Baynath)
(0,MLP Gedikli)
(0,CNN Ceker)
(0,MLP Maheshwary)
(0,XGBoost Krishna)
};
\addplot [fill=blue,opacity=1.00]
coordinates {
(0,Ours)
(0,ELM Ravindran)
(95.96,XGBoost Dwivedi)
(0,CNN Dwivedi)
(0,SVM/NEAT Baynath)
(0,MLP Gedikli)
(0,CNN Ceker)
(0,MLP Maheshwary)
(0,XGBoost Krishna)
};
\addplot [fill=blue,opacity=1.00]
coordinates {
(0,Ours)
(0,ELM Ravindran)
(0,XGBoost Dwivedi)
(94.24,CNN Dwivedi)
(0,SVM/NEAT Baynath)
(0,MLP Gedikli)
(0,CNN Ceker)
(0,MLP Maheshwary)
(0,XGBoost Krishna)
};
\addplot [fill=blue,opacity=1.00]
coordinates {
(0,Ours)
(0,ELM Ravindran)
(0,XGBoost Dwivedi)
(0,CNN Dwivedi)
(95.50,SVM/NEAT Baynath)
(0,MLP Gedikli)
(0,CNN Ceker)
(0,MLP Maheshwary)
(0,XGBoost Krishna)
};
\addplot [fill=blue,opacity=1.00]
coordinates {
(0,Ours)
(0,ELM Ravindran)
(0,XGBoost Dwivedi)
(0,CNN Dwivedi)
(0,SVM/NEAT Baynath)
(94.70,MLP Gedikli)
(0,CNN Ceker)
(0,MLP Maheshwary)
(0,XGBoost Krishna)
};
\addplot [fill=blue,opacity=1.00]
coordinates {
(0,Ours)
(0,ELM Ravindran)
(0,XGBoost Dwivedi)
(0,CNN Dwivedi)
(0,SVM/NEAT Baynath)
(0,MLP Gedikli)
(84.28,CNN Ceker)
(0,MLP Maheshwary)
(0,XGBoost Krishna)
};
\addplot [fill=blue,opacity=1.00]
coordinates {
(0,Ours)
(0,ELM Ravindran)
(0,XGBoost Dwivedi)
(0,CNN Dwivedi)
(0,SVM/NEAT Baynath)
(0,MLP Gedikli)
(0,CNN Ceker)
(93.59,MLP Maheshwary)
(0,XGBoost Krishna)
};
\addplot [fill=blue,opacity=1.00]
coordinates {
(0,Ours)
(0,ELM Ravindran)
(0,XGBoost Dwivedi)
(0,CNN Dwivedi)
(0,SVM/NEAT Baynath)
(0,MLP Gedikli)
(0,CNN Ceker)
(0,MLP Maheshwary)
(93.79,XGBoost Krishna)
};
\end{axis}
\end{tikzpicture}
\caption{Comparison to previous work}\label{fig:compare}
\end{figure}
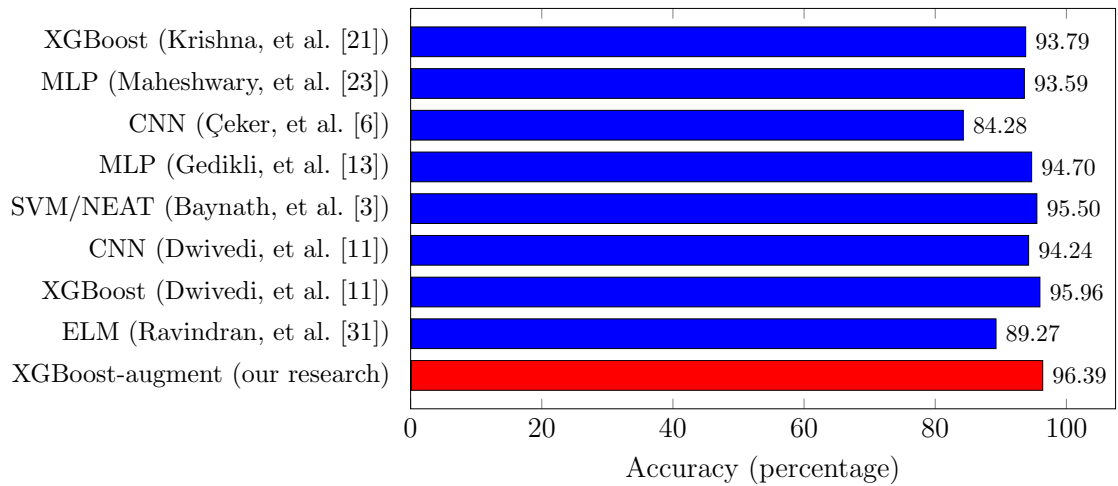

In Figure~\ref{fig:compare} we provide a comparison of
our best result to previous work. We see that our best accuracy of~96.39\%\ 
offers a modest improvement over previous work in this field.

\section{Conclusion and Future Work}
\label{Conclusion and Future Work}

In this paper, we tested and analyzed a wide variety of
machine learning techniques for biometric authentication based on 
fixed-text typing characteristics. We found that XGBoost with data augmentation
performed best, with MLP performing nearly as well. Our results improved
upon previous research involving the same dataset.

There are many avenues available for future work. For example,
we model optimization and model fusion would be interesting. 
For model optimization, we could consider
techniques from contrastive learning and self-supervised techniques
to see whether these approaches can improve our model. 

As another example of possible future work, the
robustness of various techniques can be evaluated using an 
algorithm known as POPQORN~\cite{ko2019popqorn}. The idea 
behind Popcorn is to observe the effect of outside disturbances to 
the model and thereby measure its robustness.

\bibliographystyle{plain}
\bibliography{referencesHC.bib,referencesJWcombo.bib,Stamp-Mark.bib}

\begin{thebibliography}{10}

\bibitem{2}
Faisal Alshanketi, Issa Traore, and Ahmed~Awad Ahmed.
\newblock Improving performance and usability in mobile keystroke dynamic
  biometric authentication.
\newblock In {\em 2016 IEEE Security and Privacy Workshops}, SPW, pages 66--73,
  2016.

\bibitem{8704135}
Purvashi Baynath, K.~M.~Sunjiv Soyjaudah, and Maleika Heenaye-Mamode Khan.
\newblock Machine learning algorithm on keystroke dynamics pattern.
\newblock In {\em 2018 IEEE Conference on Systems, Process and Control}, ICSPC,
  pages 11--16, 2018.

\bibitem{8883621}
Purvashi Baynath, K.~M.~Sunjiv Soyjaudah, and Maleika Heenaye-Mamode Khan.
\newblock Machine learning algorithm on keystroke dynamics fused pattern in
  biometrics.
\newblock In {\em 2019 Conference on Next Generation Computing Applications},
  {NextComp}, pages 1--6, 2019.

\bibitem{62613}
Saleh~Ali Bleha, Charles Slivinsky, and B.~Hussien.
\newblock Computer-access security systems using keystroke dynamics.
\newblock {\em IEEE Transactions on Pattern Analysis and Machine Intelligence},
  12(12):1217--1222, 1990.

\bibitem{10.1006/imms.1993.1092}
Marcus Brown and Samuel~Joe Rogers.
\newblock User identification via keystroke characteristics of typed names
  using neural networks.
\newblock {\em International Journal of Man-Machine Studies}, 39(6):999--1014,
  1993.

\bibitem{8267667}
Hayreddin \c{C}eker and Shambhu Upadhyaya.
\newblock Sensitivity analysis in keystroke dynamics using convolutional neural
  networks.
\newblock In {\em 2017 IEEE Workshop on Information Forensics and Security},
  WIFS, pages 1--6, 2017.

\bibitem{1431505}
Wendy Chen and Weide Chang.
\newblock Applying hidden {M}arkov models to keystroke pattern analysis for
  password verification.
\newblock In {\em Proceedings of the 2004 IEEE International Conference on
  Information Reuse and Integration}, IRI, pages 467--474, 2004.

\bibitem{1}
Lucian Constantin.
\newblock \textit{PC World}: {AI}-based typing biometrics might be
  authentication's next big thing.
\newblock
  \url{https://www.pcworld.com/article/3162010/ai-based-typing-biometrics-might-be-authentications-next-big-thing.html},
  2017.

\bibitem{Belief}
Yunbin Deng and Yu~Zhong.
\newblock Keystroke dynamics user authentication based on {Gaussian} mixture
  model and deep belief nets.
\newblock \url{https://www.hindawi.com/journals/isrn/2013/565183/}, 2013.

\bibitem{10.1007/1-4020-8143-X_18}
Paul~S. Dowland and Steven~M. Furnell.
\newblock A long-term trial of keystroke profiling using digraph, trigraph and
  keyword latencies.
\newblock In Yves Deswarte, Fr{\'e}d{\'e}ric Cuppens, Sushil Jajodia, and
  Lingyu Wang, editors, {\em Security and Protection in Information Processing
  Systems}, pages 275--289. Springer, 2004.

\bibitem{8628852}
Chaitanya Dwivedi, Divyanshu Kalra, Divesh Naidu, and Swati Aggarwal.
\newblock Keystroke dynamics based biometric authentication: A hybrid
  classifier approach.
\newblock In {\em 2018 IEEE Symposium Series on Computational Intelligence},
  SSCI, pages 266--273, 2018.

\bibitem{forsen1977personal}
George~E. Forsen, Mark~R. Nelson, and Raymond J.~Staron (Jr.).
\newblock Personal attributes authentication techniques.
\newblock \url{https://books.google.com.tw/books?id=tbs4OAAACAAJ}, 1977.

\bibitem{Gedikli}
Ahmet Gedikli and Mehmet Efe.
\newblock A simple authentication method with multilayer feedforward neural
  network using keystroke dynamics.
\newblock \url{https://web.cs.hacettepe.edu.tr/~onderefe/PDF/2019medprai2.pdf},
  2020.

\bibitem{3}
Ahmet~Melih Gedikli and Mehmet~{\"O}nder Efe.
\newblock A simple authentication method with multilayer feedforward neural
  network using keystroke dynamics.
\newblock In Chawki Djeddi, Akhtar Jamil, and Imran Siddiqi, editors, {\em
  Pattern Recognition and Artificial Intelligence}, pages 9--23. Springer,
  2020.

\bibitem{886039}
S.~{Haider}, A.~{Abbas}, and A.~K. {Zaidi}.
\newblock A multi-technique approach for user identification through keystroke
  dynamics.
\newblock In {\em 2000 IEEE International Conference on Systems, Man and
  Cybernetics}, SMC 2000, pages 1336--1341, 2000.

\bibitem{4656564}
Danoush Hosseinzadeh and Sridhar Krishnan.
\newblock Gaussian mixture modeling of keystroke patterns for biometric
  applications.
\newblock {\em IEEE Transactions on Systems, Man, and Cybernetics, Part C
  (Applications and Reviews)}, 38(6):816--826, 2008.

\bibitem{10.1007/978-3-540-74549-5_125}
Pilsung Kang, Seong-seob Hwang, and Sungzoon Cho.
\newblock Continual retraining of keystroke dynamics based authenticator.
\newblock In Seong-Whan Lee and Stan~Z. Li, editors, {\em Advances in
  Biometrics}, pages 1203--1211. Springer, 2007.

\bibitem{13}
Yoon Kim.
\newblock Convolutional neural networks for sentence classification.
\newblock \url{https://arxiv.org/abs/1408.5882}, 2014.

\bibitem{ko2019popqorn}
Ching-Yun Ko, Zhaoyang Lyu, Tsui-Wei Weng, Luca Daniel, Ngai Wong, and Dahua
  Lin.
\newblock {POPQORN}: Quantifying robustness of recurrent neural networks.
\newblock In {\em Proceedings of the 36th International Conference on Machine
  Learning}, pages 3468--3477, 2019.

\bibitem{6845213}
Marc~Alexander Kowtko.
\newblock Biometric authentication for older adults.
\newblock In {\em IEEE Long Island Systems, Applications and Technology
  Conference 2014}, LISAT, pages 1--6, 2014.

\bibitem{e}
Gutha~Jaya Krishna, Harshal Jaiswal, P.~Sai~Ravi Teja, and Vadlamani Ravi.
\newblock Keystroke based user identification with {XGBoost}.
\newblock In {\em 2019 IEEE Region 10 Conference}, TENCON 2019, pages
  1369--1374, 2019.

\bibitem{611659}
Daw-Tung Lin.
\newblock Computer-access authentication with neural network based keystroke
  identity verification.
\newblock In {\em Proceedings of International Conference on Neural Networks},
  ICNN'97, pages 174--178, 1997.

\bibitem{7}
Saket Maheshwary, Soumyajit Ganguly, and Vikram Pudi.
\newblock {Deep Secure}: A fast and simple neural network based approach for
  user authentication and identification via keystroke dynamics.
\newblock
  \url{http://iwaise.it.nuigalway.ie/wp-content/uploads/2017/02/DeepSecure.pdf},
  2017.

\bibitem{5}
Roy~A. {Maxion} and Kevin~S. {Killourhy}.
\newblock Comparing anomaly-detection algorithms for keystroke dynamics.
\newblock In {\em 2009 IEEE/IFIP International Conference on Dependable Systems
  \&\ Networks}, DSN, pages 125--134, 2009.

\bibitem{5544311}
Roy~A. {Maxion} and Kevin~S. {Killourhy}.
\newblock Keystroke biometrics with number-pad input.
\newblock In {\em 2010 IEEE/IFIP International Conference on Dependable Systems
  \&\ Networks}, DSN, pages 201--210, 2010.

\bibitem{7477228}
Soumik Mondal and Patrick Bours.
\newblock Combining keystroke and mouse dynamics for continuous user
  authentication and identification.
\newblock In {\em 2016 IEEE International Conference on Identity, Security and
  Behavior Analysis}, ISBA, pages 1--8, 2016.

\bibitem{Monrose}
Fabian Monrose and Aviel Rubin.
\newblock Authentication via keystroke dynamics.
\newblock In {\em Proceedings of the 4th ACM Conference on Computer and
  Communications Security}, pages 48--56, 1997.

\bibitem{MONROSE2000351}
Fabian Monrose and Aviel~D. Rubin.
\newblock Keystroke dynamics as a biometric for authentication.
\newblock {\em Future Generation Computer Systems}, 16(4):351--359, 2000.

\bibitem{MULIONO2018564}
Yohan Mulionoa, Hanry Hamb, and Dion Darmawan.
\newblock Keystroke dynamic classification using machine learning for password
  authorization.
\newblock {\em Procedia Computer Science}, 135:564--569, 2018.

\bibitem{4}
Nataasha Raul, Radha Shankarmani, and Padmaja Joshi.
\newblock A comprehensive review of keystroke dynamics-based authentication
  mechanism.
\newblock In {\em International Conference on Innovative Computing and
  Communications}, pages 149--162, 2020.

\bibitem{7435705}
Sriram Ravindran, Chandan Gautam, and Aruna Tiwari.
\newblock Keystroke user recognition through extreme learning machine and
  evolving cluster method.
\newblock In {\em 2015 IEEE International Conference on Computational
  Intelligence and Computing Research}, ICCIC, pages 1--5, 2015.

\bibitem{6966780}
Joseph Roth, Xiaoming Liu, Arun Ross, and Dimitris Metaxas.
\newblock Investigating the discriminative power of keystroke sound.
\newblock {\em IEEE Transactions on Information Forensics and Security},
  10(2):333--345, 2015.

\bibitem{Stamp04arevealing}
Mark Stamp.
\newblock A revealing introduction to hidden {M}arkov models.
\newblock \url{https://www.cs.sjsu.edu/~stamp/RUA/HMM.pdf}, 2004.

\bibitem{StampML2017}
Mark Stamp.
\newblock {\em Introduction to Machine Learning with Applications in
  Information Security}.
\newblock Chapman and Hall/CRC, Boca Raton, 2017.

\bibitem{6}
Laurens van~der Maaten and Geoffrey Hinton.
\newblock Visualizing data using {t-SNE}.
\newblock {\em Journal of Machine Learning Research}, 9(86):2579--2605, 2008.

\bibitem{1223761}
Enzhe Yu and Sungzoon Cho.
\newblock {GA-SVM} wrapper approach for feature subset selection in keystroke
  dynamics identity verification.
\newblock In {\em Proceedings of the International Joint Conference on Neural
  Networks}, pages 2253--2257, 2003.

\bibitem{5634492}
Robert~S. Zack, Charles~C. Tappert, and Sung-Hyuk Cha.
\newblock Performance of a long-text-input keystroke biometric authentication
  system using an improved $k$-nearest-neighbor classification method.
\newblock In {\em 2010 Fourth IEEE International Conference on Biometrics:
  Theory, Applications and Systems}, BTAS, pages 1--6, 2010.

\bibitem{Zhong}
Yu~Zhong and Yunbin Deng.
\newblock A survey on keystroke dynamics biometrics: Approaches, advances, and
  evaluations.
\newblock
  \url{https://sciencegatepub.com/books/gcsr/gcsr_vol2/GCSR_Vol2_Ch1.pdf},
  2015.

\end{thebibliography}

\end{document}